\documentclass[10pt,twocolumn,letterpaper]{article}

\usepackage{wacv}
\usepackage{times}
\usepackage{epsfig}
\usepackage{graphicx}
\usepackage{amsmath}
\usepackage{amssymb}
\usepackage{booktabs}
\graphicspath{./images/}
\usepackage{float}
\usepackage{subfigure}
\usepackage{cite}
\usepackage{multirow}
\usepackage[accsupp]{axessibility}  

\def\wacvPaperID{960} 

\wacvalgorithmstrack   

\wacvfinalcopy 

\ifwacvfinal
\usepackage[breaklinks=true,bookmarks=false]{hyperref}
\else
\usepackage[pagebackref=true,breaklinks=true,colorlinks,bookmarks=false]{hyperref}
\fi

\newcommand\approach{PanopticGAN}

\begin{document}

\title{Panoptic-aware Image-to-Image Translation}

\author{Liyun Zhang$^{1}$, Photchara Ratsamee$^{1,2}$, Bowen Wang$^{1}$, Zhaojie Luo$^{1}$, Yuki Uranishi$^{1}$,\\
Manabu Higashida$^{1}$ and Haruo Takemura$^{1}$\\
$^{1}$Osaka University, Japan {\tt\small liyun.zhang@lab.ime.cmc.osaka-u.ac.jp}\\
$^{2}$Osaka Institute of Technology, Japan {\tt\small photchara@ime.cmc.osaka-u.ac.jp}\\
}

\maketitle
\thispagestyle{empty}

\begin{abstract}
Despite remarkable progress in image translation, the complex scene with multiple discrepant objects remains a challenging problem.
The translated images have low fidelity and tiny objects in fewer details causing unsatisfactory performance in object recognition.
Without thorough object perception (\ie, bounding boxes, categories, and masks) of images as prior knowledge, the style transformation of each object will be difficult to track in translation.
We propose panoptic-aware generative adversarial networks (\approach) for image-to-image translation together with a compact panoptic segmentation dataset.
The panoptic perception (\ie, foreground instances and background semantics of the image scene) is extracted to achieve alignment between object content codes of the input domain and panoptic-level style codes sampled from the target style space, then refined by a proposed feature masking module for sharping object boundaries.
The image-level combination between content and sampled style codes is also merged for higher fidelity image generation.
Our proposed method was systematically compared with different competing methods and obtained significant improvement in both image quality and object recognition performance.
\end{abstract}

\begin{figure}[t]
\begin{center}
\subfigure[Image-level: BicycleGAN (Baseline)]{
\label{pipeline_a}
\includegraphics[width=1.0\linewidth]{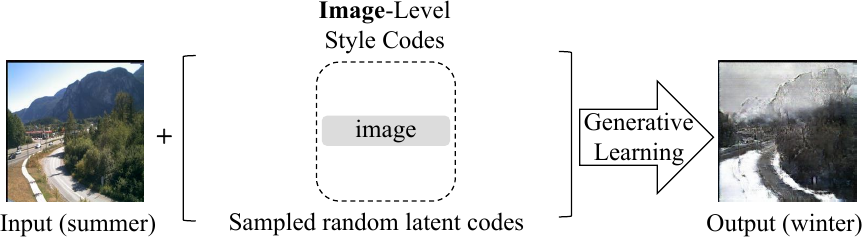}} \\
\vspace{-2.0mm}
\subfigure[Instance-level: INIT (Baseline)]{
\label{pipeline_b}
\includegraphics[width=1.0\linewidth]{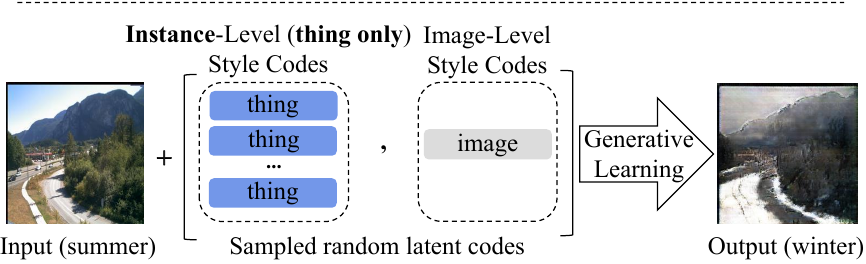}} \\
\vspace{-2.0mm}
\subfigure[Panoptic-level: \approach\ (Proposed)]{
\label{pipeline_c}
\includegraphics[width=1.0\linewidth]{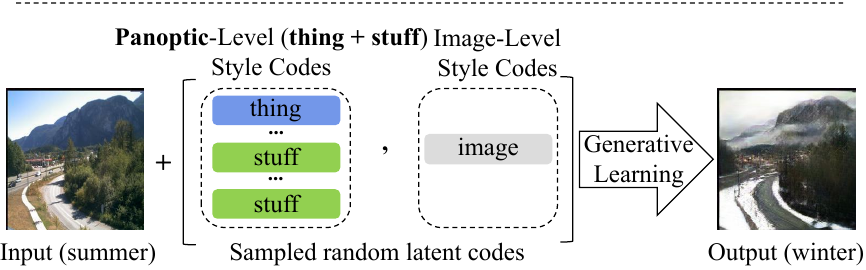}}
\end{center}
\vspace{-1.0mm}
    \caption{Pipeline comparisons of image-level \cite{BicycleGAN}, instance-level \cite{INIT} and proposed I2I translation methods. \protect\subref{pipeline_a} uses randomly sampled image-level style codes for I2I translation; \protect\subref{pipeline_b} uses instance-level (objects are only countable foreground instances `thing', \eg, car) and image-level style codes; Our approach \protect\subref{pipeline_c} uses panoptic-level (objects are both `thing' and uncountable background segments `stuff', \eg, road) and image-level style codes.}
\label{fig:pipeline}
\end{figure}

\section{Introduction}
Image-to-image (I2I) translation is a challenging problem in the computer vision field.
It needs to combine the content information of input domain image and the style of target domain \cite{MUNIT}.
Initially, some image-level I2I translation models were proposed based on paired (\eg, Pix2Pix \cite{pix2pix} or BicycleGAN \cite{BicycleGAN}) or unpaired datasets (\eg, CycleGAN \cite{CycleGAN} or MUNIT \cite{MUNIT}) and later translating high quality (realism, sharpness and diversity) images became a hot problem (\eg, Pix2PixHD \cite{Pix2PixHD} or U-GAT-IT \cite{U-GAT-IT}).
With the development of object-driven image synthesis (\eg, SG2IM \cite{SG2IM} and Layout2IM \cite{Layout2IM} synthesizes images from scenes and layout, respectively), semantics or object instance tends to promote synthesizing image with sharper objects.
Therefore, some instance-level I2I translation methods based on object instance as a perception had been proposed, \eg, INIT \cite{INIT} achieved separate learning of instances$\verb|/|$local and whole-background$\verb|/|$global areas.
They can generate high-fidelity object instances.
However, for the complex scene with multiple discrepant objects, the above methods cannot translate images to keep high fidelity and tiny objects in more detail for both foreground and background.

As illustrated in Fig.~\ref{fig:pipeline}, the image-level I2I translation method in Fig.~\ref{fig:pipeline} \subref{pipeline_a} extracts image representation as content codes to combine with image-level style codes, which are randomly sampled from the style space of the target domain for I2I translation.
In contrast, the instance-level method in Fig.~\ref{fig:pipeline} \subref{pipeline_b} uses the pre-trained instance segmentation network \cite{Mask-RCNN} to extract instance perception from the input image, it provides bounding boxes, categories, and masks of `thing' (foreground object instances).
The `thing' representations are extracted by Region of Interest Align (RoIAlign) \cite{Mask-RCNN} via bounding boxes from the image representation and then combine with the image representation as content codes.
Sampled instance-level and image-level style codes are aligned with the corresponding content codes for an instance-aware I2I translation, which can refine the foreground object instance representation precisely but does not fully refine background semantic regions.

In this paper, our proposed panoptic-level method (\approach) in Fig.~\ref{fig:pipeline} \subref{pipeline_c} uses a pre-trained panoptic segmentation network \cite{Panoptic-FPN} to extract panoptic perception, it provides bounding boxes, categories, and masks of `thing' (foreground object instances) and `stuff' (background semantic regions).
The `thing' and `stuff' representations are extracted and then combine with image representation as whole content codes.
Sampled panoptic-level and image-level style codes are aligned with the corresponding content codes for a panoptic-aware I2I translation, which can thoroughly refine each recognizable area in the image via panoptic perception for tracking the style transformation in translation and avoid losing too much information.
Our main contributions are threefold:

\begin{itemize}
  \item {\bf A novel GAN framework for panoptic-aware I2I translation:} The proposed framework extracts panoptic perception to align object content codes of the input domain with sampled panoptic-level style codes of the target domain, the image-level combination between content and sampled style codes is also merged for panoptic-aware image translation, which has high fidelity and tiny objects in more detail.
  \item {\bf A feature masking module for sharping object boundaries:} The style-aligned feature maps are further refined by feature masking to obtain sharp object boundaries for higher fidelity image generation.
  \item {\bf A compact panoptic segmentation thermal image dataset:} We annotated a panoptic segmentation thermal image dataset on partial KAIST-MS dataset \cite{KAIST-MS}, augmented dataset can be used for training a panoptic segmentation model to extract panoptic perception of thermal images in I2I translation or other tasks.
\end{itemize}

\section{Related work}
{\bf Image-to-image translation.}
Image-to-image (I2I) translation models transform the input domain image to target domain, it changes style but keeps content unchanged.
Pix2Pix \cite{pix2pix} achieved paired dataset learning, but it generates single-modal output.
BicycleGAN \cite{BicycleGAN} achieved a bijective mapping between latent and output spaces for multi-mode results.
CycleGAN \cite{CycleGAN} uses cycle consistency loss for unpaired training.
The disentangled representation models \cite{UNIT, MUNIT, DRIT} combine input domain content and target domain style for unsupervised learning.
Pix2PixHD \cite{Pix2PixHD} can translate high-resolution images by a multi-scale discriminator and coarse2fine generator.
AGGAN \cite{AGGAN} and U-GAT-IT \cite{U-GAT-IT} extracted attention regions as guidance to localize important content for high-quality results.
TSIT \cite{TSIT} uses a two-stream model with feature transformations for coarse-to-fine image synthesis. 
However, for the complex scene with multiple discrepant objects, the above methods cannot generate images in high fidelity.

\begin{figure*}[ht]
\begin{center}
\subfigure[Training manner]{
\label{overview_a}
\includegraphics[width=0.25\linewidth]{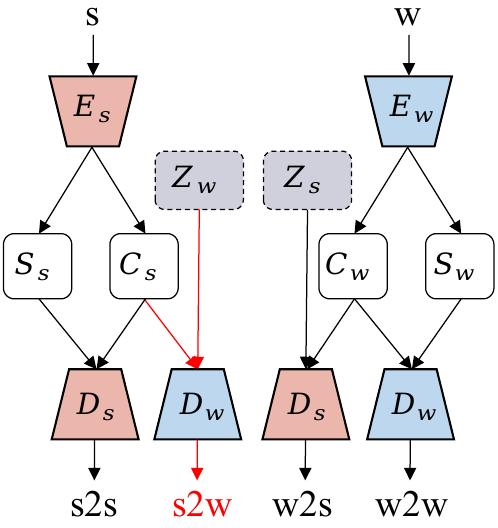}}\subfigure[Pipeline]{
\label{overview_b}
\includegraphics[width=0.75\linewidth]{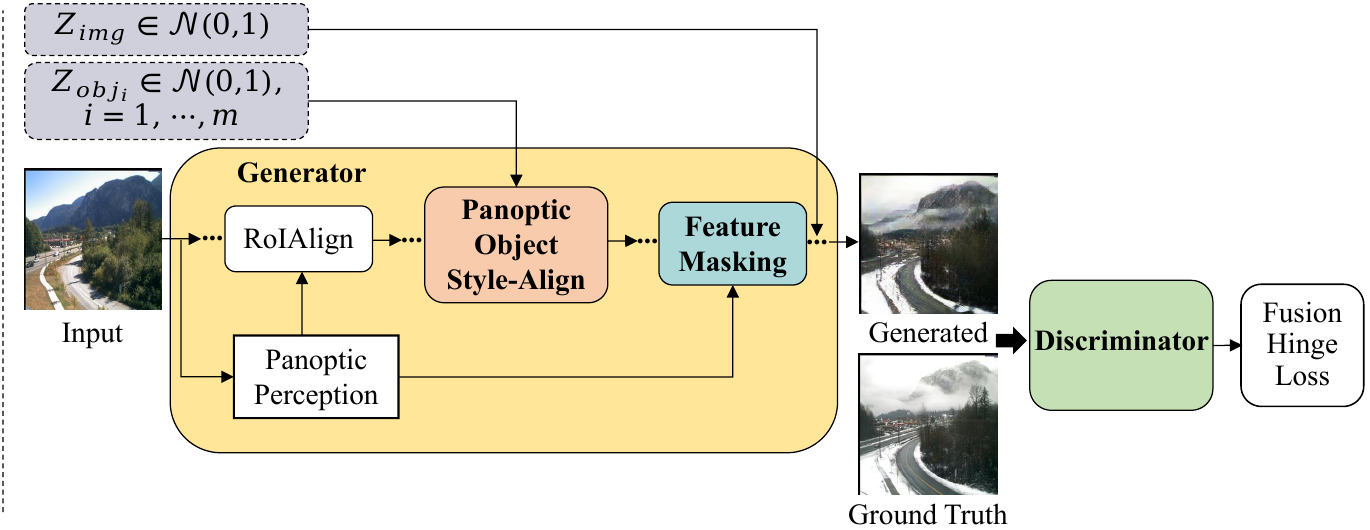}}
\end{center}
\vspace{-2.0mm}
    \caption{Illustration of the training manner and pipeline for \approach. The red arrows in \protect\subref{overview_a} corresponds to the process of \protect\subref{overview_b}.}
\label{overview}
\end{figure*}

{\bf Instance-level image-to-image translation.}
The Instance-level I2I translation is derived from the object-driven image generation methods (\eg, synthesizing images from object scenes \cite{Scene-Generation, SG2IM} or generating images from layouts \cite{Layout2IM, OC-GAN, LostGAN}), they use object perception (\ie, bounding boxes or masks) for generating sharp object boundaries.
Instagan \cite{Instagan} incorporated a set of instance attributes for instance-aware I2I translation.
DA-GAN \cite{Da-gan} learned a deep attention encoder to consequently discover instance-level correspondences.
SCGAN \cite{SCGAN} and SalG-GAN \cite{SalG-GAN} regarded saliency maps as an object perception for image translation.
Shen \etal \cite{INIT}, Su \etal \cite{Instance-colorization} and Chen \etal \cite{SentiGAN} combined the instance-level feature maps with image-level feature maps for high quality instance-level I2I translation.
However, they only use the instance-level objects `thing' without considering specific background semantic regions `stuff' in the image translation process.

\begin{figure*}[ht]
\begin{center}
\includegraphics[width=1.0\linewidth]{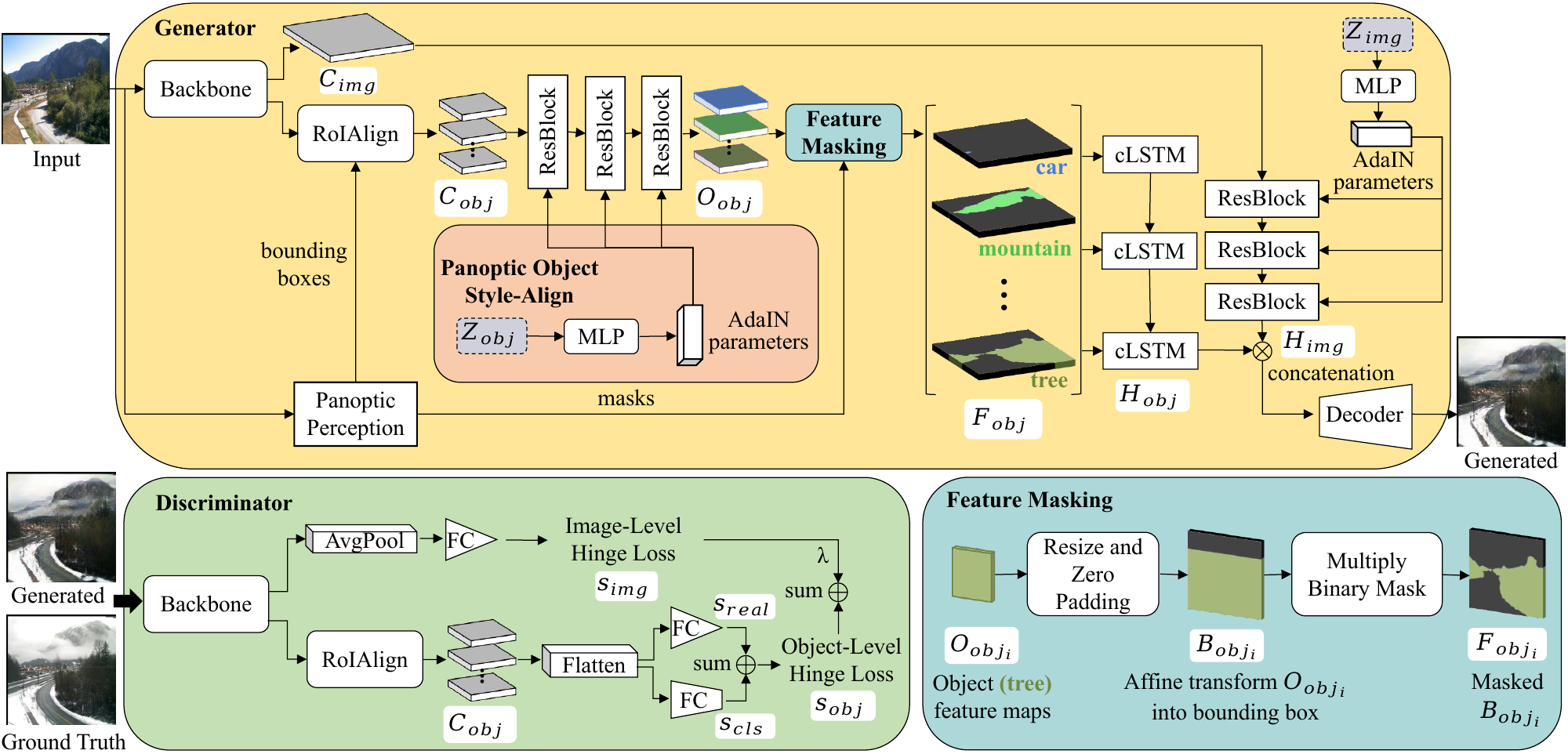}
\end{center}
\vspace{-2.0mm}
   \caption{Illustration of the architecture of our proposed \approach; the detailed notations are described in the Architecture section.}
\label{fig:architecture}
\end{figure*}

{\bf Panoptic-level image-to-image translation.}
To the best of our knowledge, the panoptic-level I2I translation problem has not yet been investigated.
From the theoretical perspective, instance-level I2I translation only considers foreground instances as objects for learning, it has certain disadvantages compared with panoptic-level I2I translation, which regards both foreground `thing' and background `stuff' as objects.
Lin \etal \cite{Attention-Guidance} extracts image regions for the discriminator to improve the performance of GANs, Huang \etal \cite{Semantic-Example-Guided} controls the output based on references semantically.
Dundar \etal \cite{Panoptic-synthesis} proved panoptic perception makes generated images have higher fidelity and tiny objects in more detail.
Panoptic segmentation \cite{Panoptic-Segmentation} combines semantic segmentation and instance segmentation to define uncountable background semantics (\eg, sky) as `stuff' and countable foreground instances (\eg, car) as `thing'.
We use a pre-trained panoptic segmentation network \cite{Panoptic-FPN} to extract panoptic perception (covers `thing' and `stuff') to make sampled panoptic-level style codes and image-level style codes combine corresponding content codes for a higher fidelity panoptic-aware I2I translation.

\section{The proposed method}
\subsection{Overview}
We provide an overview of our proposed method from the training manner and pipeline, using the summer-to-winter (transforming summer domain to winter domain) I2I translation as an example describes our framework details.

{\bf Training manner.}
In Fig.~\ref{overview} \subref{overview_a}, we use summer image $s$ and winter image $w$ from the Transient Attributes dataset \cite{Transient-attributes} to extract content codes (summer: $C_s$, winter: $C_w$) and style codes (summer: $S_s$, winter: $S_w$).
$E_s$ and $E_w$ are encoders of $s$ and $w$; $D_s$ and $D_w$ are decoders.
By combining $C_s$ with $S_s$ to feed into $D_s$, we can reconstruct summer image $s2s$.
Similarly, $w2w$ can be reconstructed by $C_w$ with $S_w$.
Style codes $Z_{w}$ and $Z_{s}$ are randomly sampled from a normal distribution.
By hypothesizing that $Z_{w}$ is from the winter style space and combining $C_s$ with $Z_{w}$ to feed into $D_w$, it can synthesize winter image $s2w$ as indicated by the red arrows.
Similarly, $w2s$ can be translated by $C_w$ and $Z_{s}$.
The cross-domain ($s2w$ and $w2s$) and within-domain ($s2s$ and $w2w$) are trained together \cite{MUNIT}.

{\bf Pipeline.}
In Fig.~\ref{overview} \subref{overview_b}, we use a pre-trained panoptic segmentation network to obtain the panoptic perception of the input image scene, it provides panoptic-level bounding boxes, categories, and masks.
The bounding boxes are provided to RoIAlign \cite{Mask-RCNN} and masks are provided to the proposed feature masking module.
Firstly, Image-level representation is extracted, panoptic-level style codes $Z_{obj}$ and image-level style codes $Z_{img}$ are sampled from normal distribution, $Z_{obj} = \left\{Z_{{obj}_i}\right\}_{i=1}^{m}$ is processed by the proposed panoptic object style-align module, $m$ is the number of objects perceived in the panoptic perception.
Note that we treat both `thing' and `stuff' as objects in panoptic perception.
$Z_{obj}$ and $Z_{img}$ will be aligned with corresponding panoptic-level and Image-level representations in the generator for panoptic-level image translation.
The translated images are fed into the discriminator, where we use fusion hinge loss consisting of image-level and object-level adversarial hinge loss terms \cite{Geometric-gan} for optimization.

\subsection{Architecture}
Our architecture, as illustrated in Fig.~\ref{fig:architecture}, is built upon a generator, discriminator and proposed novel modules (panoptic object style-align and feature masking).
We deploy a generative adversarial learning setting via summer and winter domain images from the Transient Attributes dataset \cite{Transient-attributes} for illustration of our architecture.

\subsubsection{Generator}
In generator, the input summer image $s$ (\eg, $256\times256$) is extracted by a backbone module consisting of down-sampling residual blocks for obtaining image content codes $C_{img}$ (size $32\times32$, dimension $256$).
Let $P = \left\{({category}_i,{bbox}_i,{mask}_i)_{i=1}^{m}\right\}$ be panoptic perception consisting of categories, bounding boxes, and masks, where $m$ is the number of objects perceived from a pre-trained panoptic segmentation network and ${category}_i \in CAT$ ($CAT$ defines 134 categories in the COCO-Panoptic dataset \cite{Panoptic-Segmentation}, here `thing' has 80 categories and `stuff' has 54 categories).
$C_{img}$ is cropped by RoIAlign \cite{Mask-RCNN} through object bounding boxes of $P({bbox}_i)_{i=1}^{m}$ into object content codes $C_{obj} = \left\{C_{{obj}_i}\right\}_{i=1}^{m}$ (size $8\times8$, dimension $128$).
Define $Z_{img}$ as image-level style codes (dimension $256$) and $Z_{obj} = \left\{Z_{{obj}_i}\right\}_{i=1}^{m}$ as panoptic-level style codes (dimension $64$), which are randomly sampled from normal distribution.
The goal of the generator in the summer-to-winter translation is to learn a generation function $G(\cdot)$, which is capable of translating summer image $s$ to a generated winter image $w^{'}$ via a given $(Z_{img}, Z_{obj})$:

\begin{equation}
w^{'} = G(s|Z_{img}, Z_{obj};\Theta_{G})
\end{equation}

where $\Theta_{G}$ are the parameters of the generation function.

{\bf Panoptic object style-align.}
We use a MLP network to process $Z_{obj}$ to dynamically generate the parameters $y = (y_{\gamma}, y_{\beta})$ of Adaptive Instance Normalization (AdaIN) \cite{StyleGAN} layers, then $C_{obj}$ are processed by the residual blocks with AdaIN layers.
The parameters of AdaIN layers fuse panoptic-level style with content to translate the different objects in the target image.

\begin{equation}
AdaIN(x_i, y) = y_{\gamma, i}\left(\dfrac{x_i - \mu(x_i)}{\sigma(x_i)}\right) + y_{\beta, i}
\end{equation}

where $x_i$ is each feature map of $C_{obj}$, which is normalized separately and then scaled and biased using the corresponding scalar components from style $y$.
The $\mu$ and $\sigma$ are channel-wise mean and standard deviation, $\gamma$ and $\beta$ are AdaIN parameters generated from $Z_{obj}$.
This process achieves panoptic object style-align, we obtained the style-aligned object representation $O_{obj} = \left\{O_{{obj}_i}\right\}_{i=1}^{m}$.

\begin{equation}
O_{obj} = AdaIN(C_{obj}, Z_{obj})
\end{equation}

Similarly, image-level style codes $Z_{img}$ are also processed by a MLP network to generate AdaIN parameters, which fuse the image-level style with image content codes $C_{img}$ by the residual blocks with AdaIN layers to obtain a hidden representation $H_{img}$.

{\bf Feature masking.}
As illustrated in Fig.~\ref{fig:architecture}, $O_{obj}$ contains $m$ object feature maps $\left\{O_{{obj}_i}\right\}_{i=1}^{m}$.
Since the object bounding boxes $P({bbox}_i)_{i=1}^{m}$ define the size and location of each object in the original image, we firstly affine transform each object feature maps $O_{{obj}_i}$ into its corresponding original bounding box, secondly we do zero padding outside each bounding box in the image to obtain new object feature maps $B_{obj} = \left\{B_{{obj}_i}\right\}_{i=1}^{m}$.
To remove the redundant background information outside the object contour, we further refine $B_{obj}$ via object masks $M = P({mask}_i)_{i=1}^{m}$ for more precise object boundaries.
Compared with the Convolutional Feature Masking (CFM) layer \cite{CFM} using the pixel projection method, after affine transformation the size of each feature map in $B_{obj}$ is the same as masks M, therefore we only need to align $B_{obj}$ and M along the category sequence of $1 \sim m$ and multiply to mask the values outside of object contour.
Finally, we can obtain finer object feature maps $F_{obj} = \left\{F_{{obj}_i}\right\}_{i=1}^{m}$.

\begin{equation}
F_{obj} = B_{obj} \cdot M
\end{equation}

We feed $F_{obj}$ into three layers convolutional Long-Short-Term Memory (cLSTM) module (see supplementary material) to integrate each object feature maps $\left\{F_{{obj}_i}\right\}_{i=1}^{m}$ along object sequence of $1 \sim m$ to obtain fused hidden representation $H_{obj}$.
We concatenate $H_{obj}$ with $H_{img}$ as $H$, which is up-sampled by the decoder consisting of up-sampling residual blocks to generate translated winter image $w^{'}$.

\subsubsection{Discriminator}
As illustrated in Fig.~\ref{fig:architecture}, our discriminator consists of image-level and object-level classifiers.
Similar to generator, the translated image is encoded by the backbone as image content codes $C_{img}$, which is refined by RoIAlign \cite{Mask-RCNN} as object content codes $C_{obj} = \left\{C_{{obj}_i}\right\}_{i=1}^{m}$ via bounding boxes $P({bbox}_i)_{i=1}^{m}$.
The image-level classifier consists of a global average pooling and one-output fully connected (FC) layer to process $C_{img}$ to obtain a scalar realness score $s_{img}$.
The object-level classifier consists of a flatten layer and two FC layers.
One FC layer processes $C_{obj}$ to compute a realness score for each object, denoted by $s_{real} = \left\{s_{{real}_i}\right\}_{i=1}^{m}$.
Another FC layer computes a category projection score \cite{BigGANs, cGANs, LostGAN} for each object, denoted by $s_{cls} = \left\{s_{{cls}_i}\right\}_{i=1}^{m}$, which is the inner product between category embedding (transforming each category of $P({category}_i)_{i=1}^{m}$ to a corresponding latent vector sampled from normal distribution) and linear projection (using a FC layer) of down-sampled $C_{obj}$.
Therefore, the overall object-level loss of an object is $s_{{obj}_i} = s_{{real}_i} + s_{{cls}_i}$.
The discriminator will be denoted by $D(\cdot,\Theta_{D})$ with parameters $\Theta_{D}$.

\begin{equation}
(s_{img}, s_{{obj}_i}, \cdots, s_{{obj}_m}) = D(I;\Theta_{D})
\end{equation}

Given an image $I$ (ground truth $w$ or generated $w^{'}$), the discriminator computes the prediction score for the image and the average scores for objects.

\subsection{Loss function}
The full objective comprises three loss functions:

{\bf Adversarial loss.}
We utilize image-level and object-level fusion hinge version \cite{Geometric-gan} of standard adversarial loss \cite{GANs} to train $(\Theta_{G},\Theta_{D})$ in our \approach,

\begin{small}
\begin{equation}
l_k(I)=\left\{
\begin{aligned}
& \min(0,-1 + s_k); \quad \text{if $I$ is ground truth $w$} \\
& \min(0,-1 - s_k); \quad \text{if $I$ is generated $w^{'}$} \\
\end{aligned}
\right.
\end{equation}
\end{small}

where $k \in \left\{img,{obj}_i,\cdots,{obj}_m\right\}$.
The overall loss is $l(I) = \lambda \cdot l_{img}(I) + {\dfrac{1}{m}} \begin{matrix} \sum_{i=1}^m l_{{obj}_i}(I) \end{matrix}$ with trade-off parameter $\lambda$ (1.0 used in experiment) in fusion hinge losses between image-level and object-level.
We define the losses for the discriminator and generator respectively \cite{LostGAN},

\begin{equation}
\begin{aligned}
& L_{\mathrm{adv}}(\Theta_{D},\Theta_{G})=-\underset{(I)~p_{all}(I)}{\mathbb{E}}\left[l(I)\right] \\
& L_{\mathrm{adv}}(\Theta_{G},\Theta_{D})=-\underset{(I)~p_{fake}(I)}{\mathbb{E}}\left[D(I;\Theta_{D})\right] \\
\end{aligned}
\end{equation}

where minimizing $L_{\mathrm{adv}}(\Theta_{D},\Theta_{G})$ makes discriminator to distinguish ground truth and translated images; minimizing $L_{\mathrm{adv}}(\Theta_{G},\Theta_{D})$ fools discriminator by translating fine-grained images.
$p_{all}(I)$ represents ground truth and translated images, $p_{fake}(I)$ represents translated images.

{\bf Image reconstruction loss.}
We penalize the $L_1$ difference by $L_{1}^{img} = \left \| w^{'} - w \right \|_1$ between the translated image $w^{'}$ and ground truth $w$, $\|_1$ calculates the L1 norm.
Here, we mainly calculate the within-domain ($s2s$ and $w2w$) way.

{\bf Perceptual loss.}
The $L_{\mathrm{p}}$ alleviates the problem that translated images are prone to producing distorted textures,

\begin{small}
\begin{equation}
L_{\mathrm{p}} = \begin{matrix}\sum_k\end{matrix} \dfrac{1}{C_kH_kW_k} \sum_{i=1}^{H_k} \sum_{j=1}^{W_k} \left \| \phi_{k}(w^{'})_{i,j} - \phi_{k}(w)_{i,j} \right \|_1 
\end{equation}
\end{small}

where $\phi_{k}(\cdot)$ represents feature representations of the $k$th max-pooling layer in VGG-19 network \cite{Perceptual-loss}, and $C_kH_kW_k$ represents the size of feature representations.

{\bf Full objective.}
The final loss function is defined as:

\begin{equation}
\begin{aligned}
& L_{\mathrm{total}} = \lambda_1L_{\mathrm{adv}} + \lambda_2L_{1}^{img} + \lambda_3L_{\mathrm{p}} \\
\end{aligned}
\end{equation}

where $\lambda_i$ are the parameters balancing different losses.

\subsection{Implementation details}
In $L_{\mathrm{total}}$, the $\lambda_1 \sim \lambda_3$ were set to 0.1, 1 and 10.
Model parameters were initialized using the Orthogonal Initialization method \cite{Orthogonal-Init}.
The spectral normalization \cite{SN} is to stabilize the training in both the generator and discriminator. 
We used leaky-ReLU with a slope of 0.2 for the activation function and Adam optimizer \cite{Adam} with $\beta_1 = 0$ and $\beta_2 = 0.9$.
The learning rates were set to $10^{-4}$ for the generator and $0.005$ for the discriminator.
We set 400,000 iterations for training on four NVIDIA V100 GPUs.

\begin{table*}
  \begin{center}
    {\small{
\begin{tabular}{llllllllllllr}
\toprule
\multirow{2}{*}{Method} & \multicolumn{3}{c}{HP (\%) $\uparrow$} & \multicolumn{3}{c}{IS $\uparrow$} & \multicolumn{3}{c}{FID $\downarrow$} & \multicolumn{3}{c}{DS $\uparrow$} \\
& $\text{t2c}$ & $\text{d2n}$ & $\text{s2w}$ & $\text{t2c}$ & $\text{d2n}$ & $\text{s2w}$ & $\text{t2c}$ & $\text{d2n}$ & $\text{s2w}$ & $\text{t2c}$ & $\text{d2n}$ & $\text{s2w}$ \\
\midrule
MUNIT+Seg \cite{MUNIT} & $0.8$ & $0.4$ & $0.7$ & $2.29$ & $1.50$ & $1.92$ & $98.5$ & $98.7$ & $93.9$ & $0.46$ & $0.65$ & $0.62$ \\
BicycleGAN+Seg \cite{BicycleGAN} & $3.0$ & $2.2$ & $1.7$ & $2.61$ & $1.86$ & $1.81$ & $98.8$ & $97.9$ & $92.2$ & $0.47$ & $0.60$ & $0.61$ \\
TSIT+Seg \cite{TSIT} & $12.3$ & $12.4$ & $17.1$ & $2.64$ & $1.78$ & $1.96$ & $95.3$ & $80.8$ & $81.3$ & $0.43$ & $0.67$ & $0.64$ \\
SCGAN \cite{SCGAN} & $8.4$ & $9.1$ & $8.0$ & $2.59$ & $1.62$ & $1.58$ & $96.8$ & $92.4$ & $86.4$ & $0.39$ & $0.53$ & $0.49$ \\
INIT \cite{INIT} & $34.1$ & $36.3$ & $32.0$ & $2.70$ & $1.22$ & $1.84$ & $83.2$ & $76.7$ & $78.9$ & $0.37$ & $0.65$ & $0.57$ \\
Ours & $\bf{41.2}$ & $\bf{39.4}$ & $\bf{40.1}$ & $\bf{2.85}$ & $\bf{1.93}$ & $\bf{2.01}$ & $\bf{72.7}$ & $\bf{69.4}$ & $\bf{71.1}$ & $\bf{0.54}$ & $\bf{0.72}$ & $\bf{0.69}$ \\
\bottomrule
\end{tabular}
}}
\end{center}
\vspace{-1.0mm}
\caption{Human Preference (HP), Inception Score (IS), Fréchet Inception Distance (FID) and Diversity Score (DS) metrics evaluate image quality in thermal-to-color ($\text{t2c}$), day-to-night ($\text{d2n}$) and summer-to-winter ($\text{s2w}$) tasks. Higher HP, IS and DS, and lower FID are better.}
\label{tab:qualitative}
\end{table*}

\section{Experiments}
We conducted extensive experiments to evaluate our method with state-of-the-art models to show superiority in aspects of image quality and object recognition performance.
For competing methods, MUNIT \cite{MUNIT}, BicycleGAN \cite{BicycleGAN} and TSIT \cite{TSIT} belong to image-level I2I translation.
SCGAN \cite{SCGAN} uses a saliency map as object perception for instance-level I2I translation.
INIT \cite{INIT} is an instance-level method, we also implemented it for a more fair evaluation comparison.
To achieve an adequately fair comparison, we add panoptic perception to image-level competing methods, \ie, MUNIT, BicycleGAN and TSIT.
The panoptic perception is extracted from a pre-trained panoptic segmentation network \cite{Panoptic-FPN} and concatenated with image features as an additional feature channel for training, hence we call them MUNIT+Seg, BicycleGAN+Seg and TSIT+Seg.
Based on specific metrics, we summarized the evaluation results as qualitative and quantitative aspects to discuss respectively.
Note that the model efficiency, more experimental results and limitations are provided in supplementary material.

\begin{figure*}[t]
\begin{center}
\subfigure{
\includegraphics[width=1.0\linewidth]{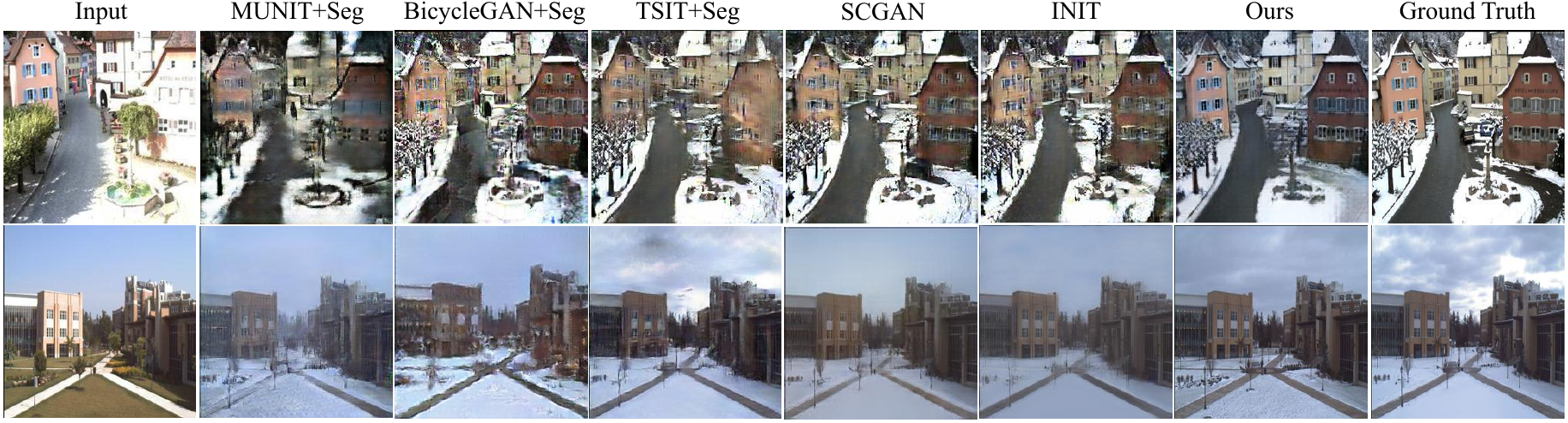}} \\
\vspace{-3.0mm}
\subfigure{
\includegraphics[width=1.0\linewidth]{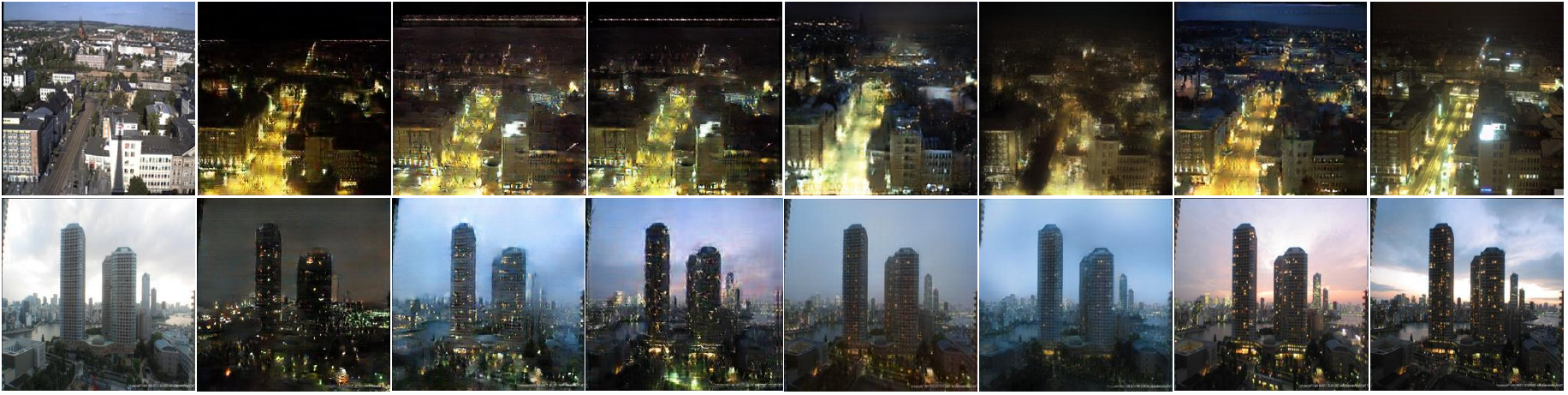}} \\
\vspace{-3.0mm}
\subfigure{
\includegraphics[width=1.0\linewidth]{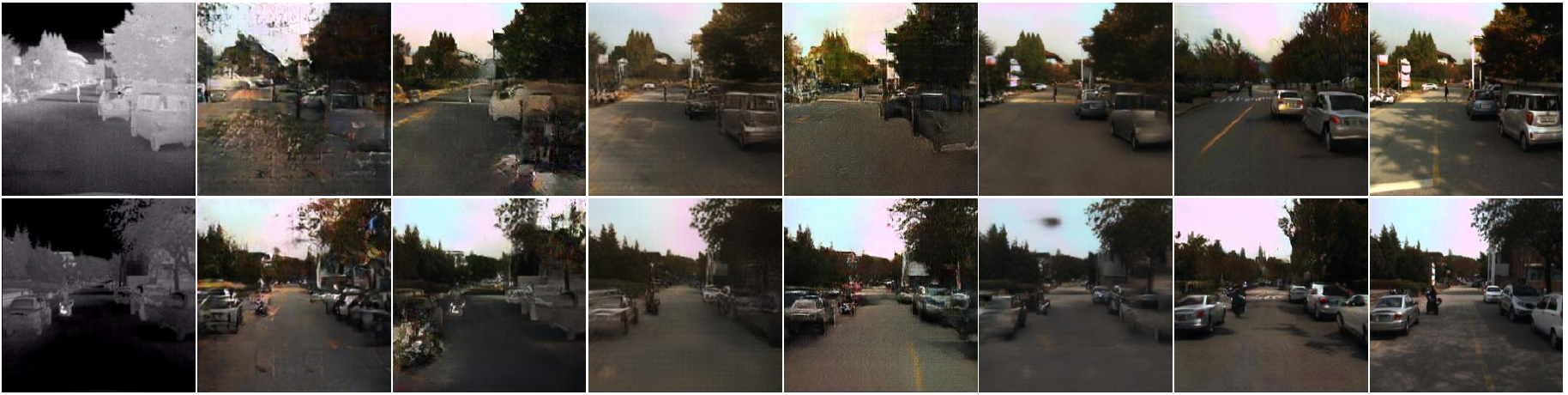}} \\
\end{center}
\vspace{-2.0mm}
    \caption{Comparison of image quality for translated images. Top: $\text{summer} \xrightarrow\ \text{winter}$; Middle: $\text{day} \xrightarrow\ \text{night}$; Bottom: $\text{thermal} \xrightarrow\ \text{color}$.}
\label{fig:evaluation}
\end{figure*}

\begin{table*}
  \begin{center}
    {\small{
\begin{tabular}{lllllllllr}
\toprule
Method & PQ $\uparrow$ & SQ $\uparrow$ & RQ $\uparrow$ & PQ$^{\text{Th}}$ $\uparrow$ & SQ$^{\text{Th}}$ $\uparrow$ & RQ$^{\text{Th}}$ $\uparrow$ & PQ$^{\text{St}}$ $\uparrow$ & SQ$^{\text{St}}$ $\uparrow$ & RQ$^{\text{St}}$ $\uparrow$ \\
\midrule
MUNIT+Seg \cite{MUNIT} & 3.3 & 12.1 & 4.2 & 0.6 & 9.6 & 0.8 & 9.0 & 17.5 & 11.3 \\
BicycleGAN+Seg \cite{BicycleGAN} & 4.3 & 16.8 & 5.5 & 0.8 & 13.1 & 1.2 & 10.9 & 23.9 & 13.6 \\
TSIT+Seg \cite{TSIT} & 6.4 & 17.2 & 8.1 & 2.1 & 13.3 & 3.3 & 13.9 & 26.4 & 15.3 \\
SCGAN \cite{SCGAN} & 5.6 & 15.2 & 7.4 & 1.7 & 11.8 & 2.6 & 13.6 & 22.5 & 17.4 \\
INIT \cite{INIT} & 7.2 & 19.6 & 9.0 & 3.1 & 15.4 & 3.9 & 16.7 & 29.1 & 20.9 \\
Ours & \bf{8.3} & \bf{22.7} & \bf{11.3} & \bf{4.2} & \bf{17.5} & \bf{5.1} & \bf{18.4} & \bf{31.0} & \bf{21.6} \\
\bottomrule
\end{tabular}
}}
\end{center}
\vspace{-1.0mm}
\caption{The PQ, SQ and RQ series metrics (higher is better) evaluate the object recognition performance of translated images.}
\label{tab:object-recognition}
\end{table*}

\subsection{Datasets}
We trained and evaluated our model on the Transient Attributes \cite{Transient-attributes} and KAIST-MS \cite{KAIST-MS} datasets for day-to-night, summer-to-winter and thermal-to-color I2I translation tasks respectively.
In the day-to-night task, we used 17,823 images for training and 2,287 images for evaluation; in summer-to-winter task, the training set is 17,674 images and evaluation set is 2558 images; in thermal-to-color, training set is 11,610 images and evaluation set is 2,541 images.
For panoptic perception in training and inference of day-to-night and summer-to-winter tasks, we use a Panoptic FPN model \cite{Panoptic-FPN} pre-trained on COCO-Panoptic dataset \cite{Panoptic-Segmentation} to perceive from input day and summer images respectively. 
For panoptic perception in the training of thermal-to-color task, we perceive it from the paired color images via pre-trained Panoptic FPN model on COCO-Panoptic dataset; in the inference, it is perceived from input images via pre-trained Panoptic FPN model on a compact our contributed dataset (see supplementary material) of thermal panoptic segmentation, the source data are the pairs of thermal and color images from partial KAIST-MS \cite{KAIST-MS} dataset.

\subsection{Evaluation metrics}
We use Human Preference (HP), Inception Score (IS) \cite{IS}, Fréchet Inception distance (FID) \cite{FID_KID} and Diversity Score (DS) metrics for image quality, and Panoptic Quality (PQ) \cite{Panoptic-Segmentation} series metrics for object recognition performance.
HP is a user perceptual study that compares the image quality of translated results from different methods including input and ground truth image, which are shown to twenty participants to select the best translated image corresponding to target domain (covering thermal-to-color, day-to-night and summer-to-winter tasks respectively).
IS is a popular metric to measure image quality generated from GANs.
FID improves IS by incorporating statistics from real images.
We used LPIPS metric \cite{LPIPS} to calculate DS, which measures the differences between two images via computing perceptual similarity \cite{Layout2IM}.
PQ contains segmentation quality (SQ) and recognition quality (RQ) \cite{Panoptic-Segmentation}, it combines mean Intersection over Union (mIoU) in SQ and average precision (AP) in RQ for a more comprehensive score than instance segmentation and object detection.
In addition, PQ$^{\text{Th}}$, SQ$^{\text{Th}}$, RQ$^{\text{Th}}$ are only used on `thing' (Th) categories; PQ$^{\text{St}}$, SQ$^{\text{St}}$, RQ$^{\text{St}}$ are only used on `stuff' (St) categories.
Detailed descriptions of metrics see supplementary material.

\begin{figure*}[t]
\begin{center}
\includegraphics[width=1.0\linewidth]{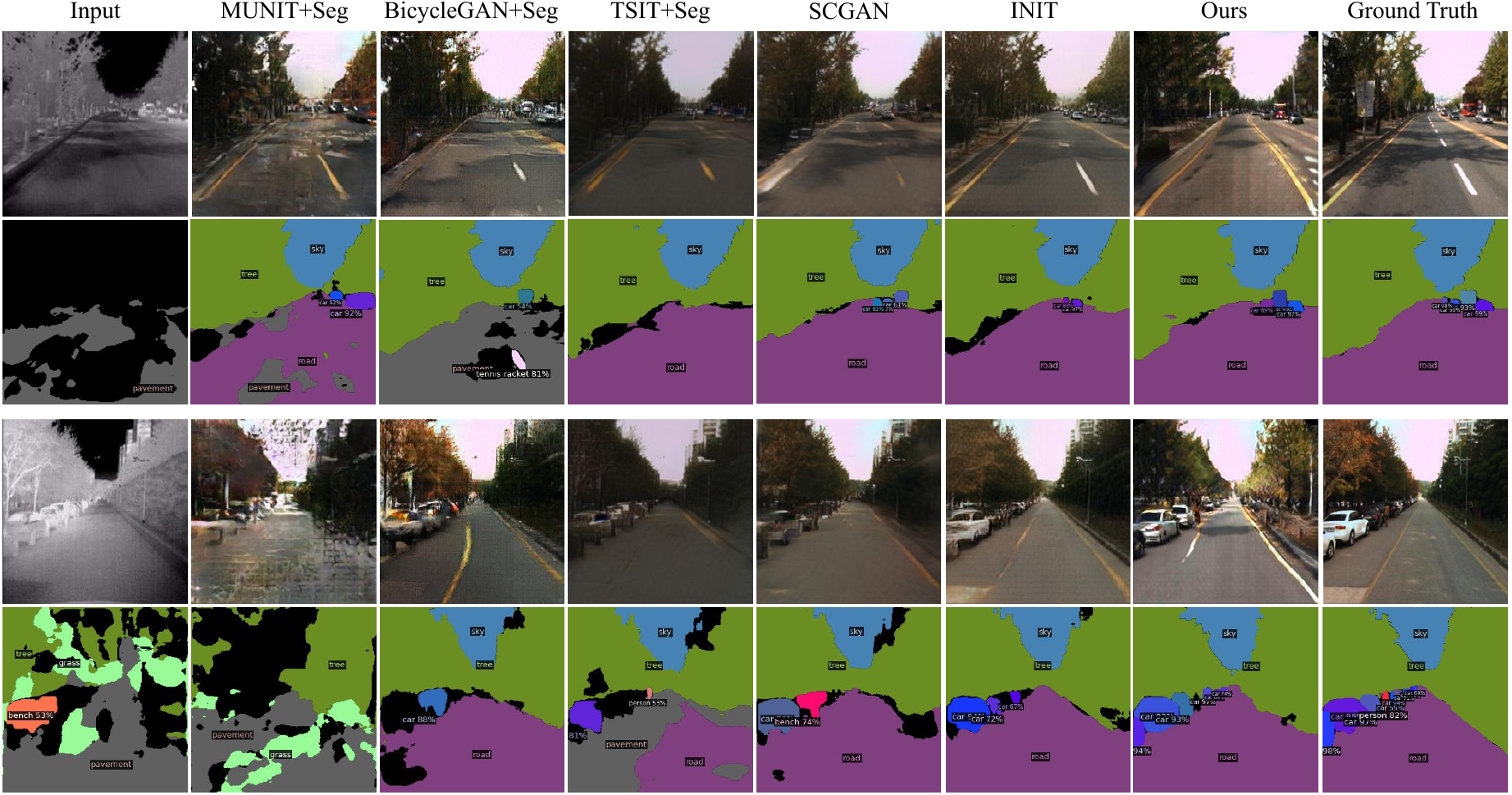}
\end{center}
\vspace{-2.0mm}
    \caption{Comparison of object recognition performance for translated images. Upper: translated images; Lower: panoptic segmentation.}
\label{fig:perception}
\end{figure*}

\subsection{Qualitative results}
For image quality, the human preference results in Table~\ref{tab:qualitative} show that our approach achieved significantly higher scores in human perceptual study of different tasks compared to other approaches.
Fig.~\ref{fig:evaluation} demonstrates our \approach\ can translate higher fidelity and brightly colored images and have tiny objects in more details.
In contrast, the results of other methods are more blurry, distorted and missing small objects.
For translated objects, our results tend to have better sharpness, more natural color style and display diversity (\eg, the appearance of cars).
On the other methods side, object sharpness is not satisfactory, the style is far from ground truth and there is insufficient diversity.

For object recognition performance, we use panoptic segmentation results by a pre-trained Panoptic FPN model on the COCO-Panoptic dataset.
We only show object recognition results on thermal-to-color task, because translated night images from day-to-night and winter images from summer-to-winter tasks have disadvantages of insignificant differences for object recognition comparison.
Fig.~\ref{fig:perception} shows that our method can achieve better object recognition performance than other methods, \eg, the number and boundaries of cars; the structure of sky, tree and road; and the areas where there are relatively fewer recognition failures.
Also, our results are significantly better than the results of original thermal images, this verified the advantages of our method when adapted for image enhancement.

\subsection{Quantitative results}
For image quality, the scores of IS, FID and DS in Table~\ref{tab:qualitative} demonstrate that our approach achieved superiority in image quality of translated images compared to other approaches.
Our method overall outperforms baselines since we avoid losing too much information in the translation.
The higher IS and lower FID of our proposed approach demonstrated that the translated images from our method have higher fidelity and sharpened object information. 
The higher DS demonstrated our method can show more flexibility and high robustness when the scene is invariant, especially for the objects generated on the image.
For object recognition performance, Table~\ref{tab:object-recognition} shows that our method performed state of the art scores compared with other competing trained models on all PQ, SQ, RQ, PQ$^{\text{Th}}$, SQ$^{\text{Th}}$, RQ$^{\text{Th}}$, PQ$^{\text{St}}$, SQ$^{\text{St}}$, RQ$^{\text{St}}$ object recognition metrics.
From the score difference, our results are uniformly higher than the state-of-the-art competing methods by a certain distance, which stated the superiority of our method.    

\subsection{Ablation study}
We demonstrated the necessity of losses and modules ($L_{\mathrm{obj}}$: object-level hinge loss; $L_{1}^{\mathrm{img}}$: image reconstruction loss; $L_{\mathrm{p}}$: perceptual loss; $M_{\mathrm{msk}}$: feature masking; $M_{\mathrm{pano}}$: panoptic object style-align; $M_{\mathrm{clstm}}$: cLSTM) of our model by comparing Inception Score (IS) \cite{IS}, Fréchet Inception Distance (FID) \cite{FID_KID} and Diversity Score (DS) \cite{Layout2IM} for image quality; Panoptic Quality (PQ) \cite{Panoptic-Segmentation}, Segmentation Quality (SQ) and Recognition Quality (RQ) are for object recognition performance.
The experiment results show several ablated versions of our model trained on the KAIST-MS \cite{KAIST-MS} dataset for the thermal-to-color task.
As shown in Table~\ref{tab:ablation}, removing any loss will decrease the overall performance.
Removing $L_{\mathrm{obj}}$ and $L_{1}^{\mathrm{img}}$ have lower IS, DS, and higher FID due to generating low fidelity image and objects with fewer variations; 
PQ, SQ, and RQ are also decreased since $L_{\mathrm{obj}}$ compute the category projection scores for objects.
Removing $L_{\mathrm{p}}$, the model produced distorted textures, this inevitably decreases image quality and object recognition performance.
Removing any $M_{\mathrm{msk}}$, $M_{\mathrm{pano}}$ or $M_{\mathrm{clstm}}$ module decreased overall performance, it demonstrates their necessity.
Because $M_{\mathrm{msk}}$ sharpens object boundaries and $M_{\mathrm{clstm}}$ sequentially integrates different objects back into image.
Especially, removing $M_{\mathrm{pano}}$ destroyed the whole foundation of our proposed panoptic-level framework, and overall performance is significantly decreased.
Therefore, the above study result for losses and modules shows the reasonability of our model design.

\begin{table}[!t]
  \begin{center}
    {\small{
\begin{tabular}{lllllll}
\toprule
Method & IS $\uparrow$ & FID $\downarrow$ & DS $\uparrow$ & PQ $\uparrow$ & SQ $\uparrow$ & RQ $\uparrow$ \\
\midrule
$\text{w/o}\ L_{\mathrm{obj}}$ & $2.24$ & $110.4$ & $0.47$ & $5.3$ & $16.2$ & $7.7$ \\
$\text{w/o}\ L_{1}^{\mathrm{img}}$ & $2.64$ & $104.3$ & $0.45$ & $6.4$ & $20.4$ & $10.1$ \\
$\text{w/o}\ L_{\mathrm{p}}$ & $2.66$ & $97.1$ & $0.42$ & $6.7$ & $19.4$ & $10.5$ \\
$\text{w/o}\ M_{\mathrm{msk}}$ & $2.53$ & $101.6$ & $0.47$ & $6.1$ & $19.8$ & $10.2$ \\
$\text{w/o}\ M_{\mathrm{pano}}$ & $2.44$ & $120.7$ & $0.43$ & $5.9$ & $18.4$ & $9.1$ \\
$\text{w/o}\ M_{\mathrm{clstm}}$ & $2.63$ & $103.2$ & $0.42$ & $6.8$ & $16.7$ & $11.1$ \\
$\text{full model}$ & $\bf{2.85}$ & $\bf{72.7}$ & $\bf{0.54}$ & $\bf{8.3}$ & $\bf{22.7}$ & $\bf{11.3}$ \\
\bottomrule
\end{tabular}
}}
\end{center}
\caption{Ablation study. The performance removing losses and modules are compared in image quality (evaluated by IS, FID and DS) and object recognition (evaluated by PQ, SQ and RQ).}
\label{tab:ablation}
\end{table}

\section{Conclusion}
We propose a novel panoptic-aware image-to-image translation method (\approach) together with a compact panoptic segmentation dataset.
The panoptic perception is extracted to achieve a panoptic-level combination between content and style codes, results are further refined by our proposed feature masking module for sharp object boundary generation.
The image-level combination of content and style codes is also merged for translating images with high fidelity and tiny objects in more detail in complex scenes with multiple discrepant objects.
Extensive experiments demonstrated that our method obtained significant improvement in image quality and object recognition performance compared to different methods.
Incorporating bounding boxes and mask regression losses in image translation model training will be the focus of our future work.

\noindent
{\bf Acknowledgments:} This work has been partly supported by the KAKENHI Fund for the Promotion of Joint International Research (fostering joint international research (B) No. 20KK0086) and the Mohamed Bin Zayed International Robotics Challenge (MBZIRC) Grant.

{\small
\bibliographystyle{ieee_fullname}
\bibliography{egbib}
}

\end{document}